\documentclass[letterpaper, 10 pt, conference]{ieeeconf} 
\IEEEoverridecommandlockouts
\usepackage{amssymb,amsmath}
\usepackage{cite} 
\usepackage{booktabs}
\usepackage[font=footnotesize,labelformat=simple]{subcaption}
\usepackage[font=footnotesize]{caption}
\usepackage[dvipsnames]{xcolor}
\usepackage{xspace}
\usepackage{amssymb,amsmath}
\usepackage{algorithm}
\usepackage[noend]{algpseudocode}
\usepackage{graphicx}
\usepackage{dsfont}

\makeatletter
\def\BState{\State\hskip-\ALG@thistlm}
\makeatother

\newcommand{\eref}[1]{(\ref{#1})}

\newcommand {\argmax}[1]{\underset{#1}{\operatorname{argmax}}}

\usepackage{hyperref}
\hypersetup{bookmarksopen,bookmarksnumbered,
pdfpagemode=UseOutlines,
colorlinks=true,
linkcolor=blue,
anchorcolor=blue,
citecolor=blue,
filecolor=blue,
menucolor=blue,
urlcolor=blue
}

\usepackage{bm}

\newcommand{\xxnote}[3]{}
\ifx\hidenotes\undefined
  \usepackage{color}
  \renewcommand{\xxnote}[3]{\color{#2}{#1: #3}}
\fi

\newcommand{\human}{\textrm{H}\xspace}
\newcommand{\robot}{\textrm{R}\xspace}

\title{
Robot Learning via Human Adversarial Games\\
}
\author{
Jiali Duan$^*$, Qian Wang$^*$, Lerrel Pinto, C.-C. Jay Kuo and Stefanos Nikolaidis%
   \thanks{$^*$ Duan and Wang contributed equally to the work. 
    }
    \thanks{Duan and Kuo are with the Department of Electrical and Computer Engineering, University of Southern California, Los Angeles 90089, USA.
        (e-mail: jialidua@usc.edu, cckuo@sipi.usc.edu).
    }
    \thanks{
        Wang and Nikolaidis are with the Department of Computer Science, University of Southern California, Los Angeles 90089, USA. 
        (e-mail: \{wang215, nikolaid\}@usc.edu).
    }
    \thanks{
        Pinto is with the Robotics Institute, Carnegie Mellon University,  Pittsburgh 15213, USA. 
        (e-mail: lerrelp@cs.cmu.edu).
    }
}

\begin{document}

\maketitle

\begin{abstract}
Much work in robotics has focused on ``human-in-the-loop'' learning techniques that improve the efficiency of the learning process. However, these algorithms have made the strong assumption of a \emph{cooperating} human supervisor that assists the robot. In reality, human observers tend to also act in an \emph{adversarial} manner towards deployed robotic systems. We show that this can in fact improve the robustness of the learned models by proposing a physical framework that leverages perturbations applied by a human adversary, guiding the robot towards  more robust models. In a manipulation task, we show that grasping success improves significantly when the robot trains with a human adversary as compared to training in a self-supervised manner. 
\end{abstract}

\section{Introduction}
\label{sec:intro}

We focus on the problem of end-to-end learning for planning and control in robotics. For instance, we want a robotic arm to learn robust manipulation grasps that can withstand perturbations using input images from an on-board camera.

Learning such models is challenging, due to the large amount of samples required. For instance, in previous work~\cite{pinto2016supersizing}, a robotic arm collected more than 50K examples to learn a grasping model in a self-supervised manner. Researchers at Google~\cite{levine2018learning} developed an arm farm and collected hundreds of thousands of examples for grasping. This shows the power of parallelizing exploration, while it requires a large amount of resources and the system is unable to distinguish between stable and unstable grasps.

To improve sample efficiency, Pinto et al.~\cite{pinto2017supervision} showed that robust grasps can be learned using a robotic adversary: a second arm that applies disturbances to the first arm. By training jointly both the first arm and the adversary, they show that this can lead to robust grasping solutions.

This configuration, however, typically requires two robotic arms placed in close proximity to each other. What if there is one robotic arm ``in the wild'' interacting with the environment, as well as with humans? 

One approach could be to have the human act as a teammate, and assist the robot in completing the task. An increasing amount of work~\cite{macglashan2017interactive, warnell2018deep, knox2009interactively, knox2012reinforcement, lin2017explore,reddy2018shared} has shown the benefits of human feedback in the robot learning process.  

At the same time, we should not always expect the human to act as a collaborator. In fact, previous studies in human-robot interaction~\cite{bartneck2008exploring,brscic2015escaping,nomura2016children} have shown that people, especially children, have acted in an adversarial and even abusive manner when interacting with robots. 

This work explores the degree to which a robotic arm could exploit such human adversarial behaviors in its learning process. Specifically, we address the following research question: 

\begin{quote}
How can we leverage human adversarial actions to improve robustness of the learned policies?
\end{quote}

\begin{figure}[!t]
\centering
\includegraphics[width=0.8\linewidth]{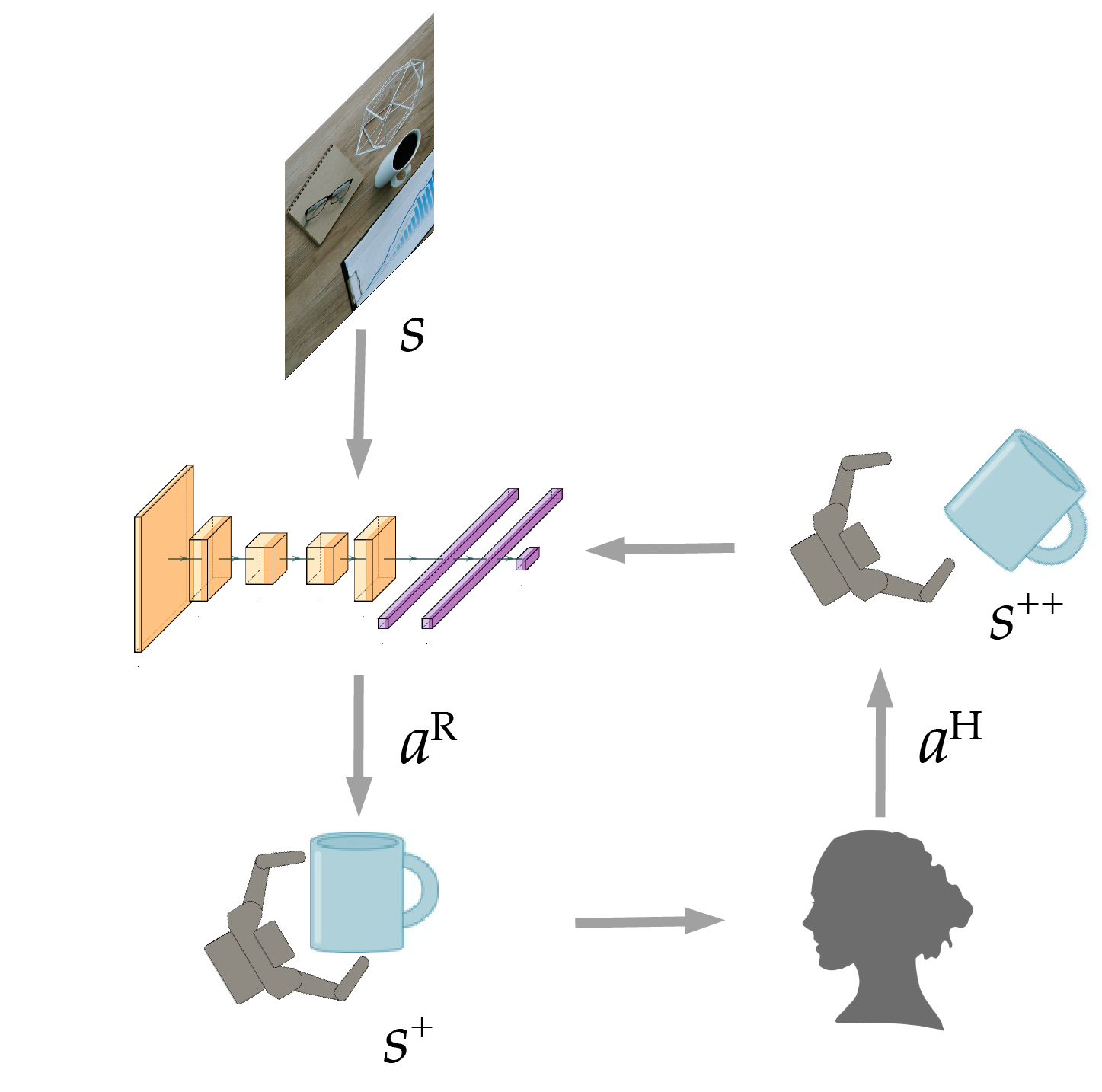}
\caption{An overview of our framework for a robot learning robust grasps by interacting with a human adversary.}
\label{fig:best}
\end{figure}

\newcommand{\linescale}{0.14}

\begin{figure*}[!t]
\centering
\begin{tabular}{ccccc}
\includegraphics[width=\linescale\linewidth]{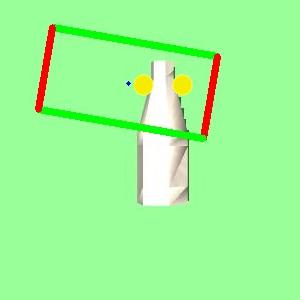}&
\includegraphics[width=\linescale\linewidth]{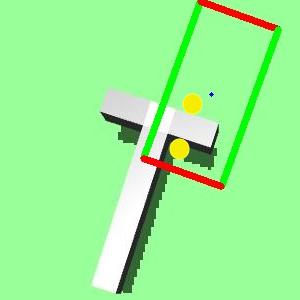}&
\includegraphics[width=\linescale\linewidth]{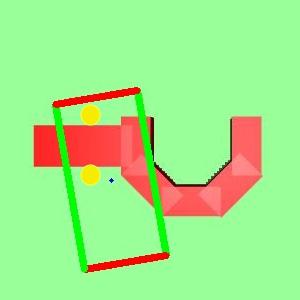}&
\includegraphics[width=\linescale\linewidth]{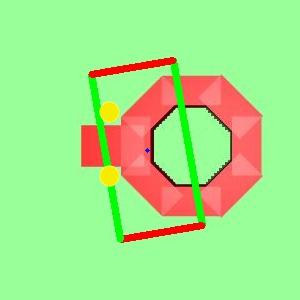}&
\includegraphics[width=\linescale\linewidth]{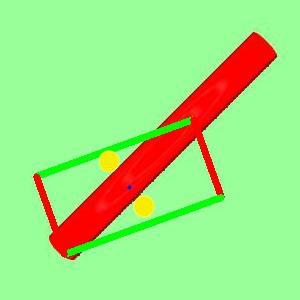}\\\includegraphics[width=\linescale\linewidth]{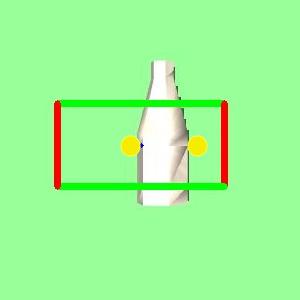}&
\includegraphics[width=\linescale\linewidth]{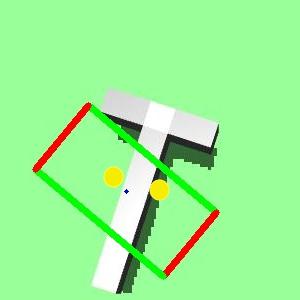}&
\includegraphics[width=\linescale\linewidth]{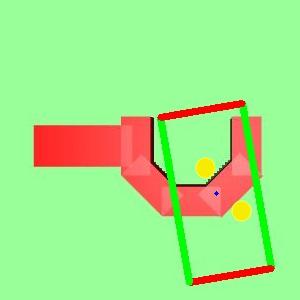}&
\includegraphics[width=\linescale\linewidth]{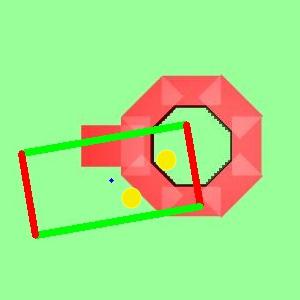}&
\includegraphics[width=\linescale\linewidth]{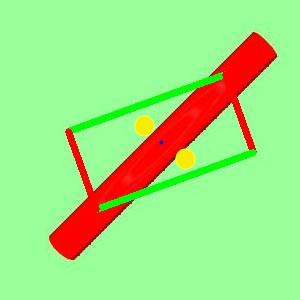}
\end{tabular}
\caption{Selected grasp predictions before (top row) and after (bottom row) training with the human adversary. The red bars show the open gripper position and orientation, while the yellow dots show the grasping points when the gripper has closed.}
\label{fig:summary}
\end{figure*}

While there has been a rich amount of human-in-the-loop learning, to the best of our knowledge this is the first effort of robot learning with adversarial human users. 
Our key insight is:
\begin{quote}
By using their domain knowledge in applying perturbations, human adversaries can contribute to the efficiency and robustness of robot learning. 
\end{quote}

We propose an ``human-adversarial'' framework where a robotic arm collects data for a manipulation task, such as grasping (Fig.~\ref{fig:best}). Instead of using humans in a collaborative manner, we propose to use them as adversaries. Specifically, we have the robot learner, and the human attempting to make the robot learner fail on its task. For instance, if the learner attempts to grasp an object, the human can apply forces to remove it from the robot. Contrary to a robot adversary in previous work~\cite{pinto2017supervision}, the human has already domain knowledge about the best way to attempt the grasp, by observing the grasp orientation and their prior knowledge of the object's geometry and physics. Additionally, here the robot can only observe one output, the outcome of the human action, rather than a distribution of adversarial actions.


We implement the framework in a virtual environment, where we allow the human to apply simulated forces on an object grasped by a robotic arm. In a user study we show that, compared to the robot learning in a self-supervised manner, the human user can provide supervision that rejects unstable robot grasps, leading to significantly more robust grasping solutions (Fig.~\ref{fig:summary}). 

While there are certain limitations on the human adversarial inputs because of the interface, this is an exciting first step towards leveraging human adversarial actions in robot learning. 
\section{Related Work}

\noindent\textbf{Self-supervised Deep Learning in Manipulation.} In robotic manipulation, deep learning has been combined with self-supervision techniques to achieve end-to-end training~\cite{lenz2015deep,levine2016end,levine2018learning}, for instance with curriculum learning~\cite{pinto2017learning}. Other approaches include learning dynamics models through interaction with objects~\cite{agrawal2016learning}. Most relevant to ours is the work by Pinto et al., where a ``protagonist'' robot learns grasping solutions by interacting with a robotic adversary. In this work, we follow a human-in-the-loop approach, where we have a robotic arm learn robust grasps by interacting with a human adversary.

\noindent\textbf{Reinforcement Learning with Human Feedback.} Previous work~\cite{knox2012reinforcement, knox2010combining,loftin2016learning,griffith2013policy,knox2008tamer,arumugam2019deep,macglashan2017interactive}  has also focused on using human feedback to augment the learning of autonomous agents.  Specifically, rather than optimizing a reward function, learning agents respond to positive and negative feedback signals provided by a human supervisor. These works have explored different ways to incorporate feedback into the learning process, either as part of the reward function of the agent, such as in the TAMER framework~\cite{knox2008tamer}, or directly in the advantage function of the algorithm, as suggested by the COACH algorithm~\cite{macglashan2017interactive}. This allows the human to train the agent towards specific behaviors, without detailed knowledge of the agent's decision making mechanism. Our work is related in that the human affects the agent's reward function. However, the human does not do this explicitly, but indirectly through its own actions. More importantly, the human acts in an adversarial manner, rather than as a collaborator or a supervisor. 

\noindent\textbf{Adversarial Methods.}  Generative adversarial methods~\cite{goodfellow2014generative, dumoulin2016adversarially} have been used to train two models, a generative model that captures the data distribution, and a discriminative model that estimates the probability that a sample came from the training data. Researchers have also analyzed a network to generate adversarial examples, with the goal of increasing the robustness of classifiers~\cite{goodfellow2014explaining}. In our case, we let a human agent generate the adversarial examples that enable adaptation of a discriminative model.

\noindent\textbf{Grasping.} We focus on generating \emph{robust} grasps, that can withstand disturbances. There is a large amount of previous work on grasping~\cite{bohg2014data,bicchi2000robotic}, that range from physics-based modeling~\cite{mahler2016dex,berenson2007grasp,berenson2008grasp} to data-driven techniques~\cite{pinto2016supersizing,levine2018learning}. The latter have focused on large-scale data collection. Pinto et al.~\cite{pinto2017supervision} have shown that perturbing grasps by shaking or snatching by a robot adversary can facilitate learning. We are interested in whether this can hold when the adversary is a human user, applying forces at the grasped object. 

\section{Problem Statement}


We formulate the problem as a \emph{two-player game with incomplete information}\cite{lavi2007algorithmic}, played by a human (\human) and a robot (\robot). We define $s \in S$ to be the \emph{state} of the world. A robot and a human are taking turns in actions.  A robot action results in a stochastic transition to new state $s^{+} \in S^{+}$, based on some unknown transition function $\mathcal{T}:S \times A^\robot \rightarrow \Pi(S^{+})$. The human then acts based on a stochastic policy, also unknown to the robot, so that $\pi^\human:(s^{+},a^\human)$. After the human and the robot's actions, the robot observes the final state $s^{++}$ and receives a reward signal $r:(s, a^\robot, s^+, a^\human, s^{++}) \mapsto r$.  

In an adversarial setting, the robot attempts to maximize $r$, while the human wishes to minimize it. Specifically, we formulate $r$ as a linear combination of two terms: the reward that the robot would receive in the absence of an adversary, and the penalty induced by the human action:  
\begin{equation}
r = R^{\robot}(s, a^\robot, s^+) - \alpha R^{\human}(s^+, a^\human, s^{++})
\label{eqn:reward}
\end{equation}

The goal of the system is to develop a policy $\pi^\robot:s \mapsto a_t^\robot$ that maximizes this reward.

\begin{equation}
\pi_*^\robot = \argmax{\pi^\robot}~ 
 {\mathds{E}} \left[r(s, a^\robot, a^\human) \vert \pi^\human \right]
\label{eqn:robot-policy}
\end{equation}

Through this maximization, the robot implicitly attempts to minimize the reward of the human adversary. In Eq. \eref{eqn:reward}, $\alpha$ controls the proportion of learning from the human's adversarial actions.

\section{Approach}
\noindent\textbf{Algorithm.} We assume that the robot's policy $\pi^\robot$ is parameterized by a set of parameters $W$, represented by a convolutional neural network. The robot uses its sensors to receive a state representation $s$, and samples an action $a^\robot$. It then observes a new state $s^{+}$, and waits for the human adversary to act. Finally, it observes the final state $s^{++}$, and computes the reward $r$ based on Eq.~\eref{eqn:reward}. A new world state is then sampled randomly, as the robot attempts to grasp a potentially different object (Algorithm~\ref{alg:algorithm}).

\noindent\textbf{Initialization.} We initialize the parameters $W$ by optimizing only for $R^{\robot}(s, a^\robot, s+)$, that is for the reward in the absence of the adversary. This allows the robot to choose actions that have a high probability of grasp success, which in turn enables the human to act in response. After training in a self-supervised manner, the network can be refined through interactions with the human.

\begin{algorithm}
\caption{Learning with a Human Adversary}\label{euclid}
\begin{algorithmic}[1]
\State Initialize parameters $W$ of robot's policy $\pi^\robot$
\For {batch $=1,B$}
\For {episode $=1,M$}
\State observe $s$ 
\State sample action $a^\robot \sim  \pi_*^\robot(s)$
\State execute action $a^\robot$ and observe $s^+$
\If{$s^+$ is not terminal}
\State observe human action $a^\human$ and state $s^{++}$
\EndIf
\State observe $r$ given by Eq.~\eref{eqn:reward} 
\State record $s, a^\robot, r$
\EndFor
\State update $W$ based on recorded sequence
\EndFor
\State return $W$
\end{algorithmic}
\label{alg:algorithm}
\end{algorithm}

\section{Learning Robust Grasps}
We instantiate the problem in a grasping framework. The robot attempts to grasp an object. The human observes the robot's grasp. If the grasp is successful, the human can apply a force as a disturbance in the robot's hand, in six different directions. In this work, we use a \emph{simulation} environment to simulate the grasps and interactions with the human. We use this environment as a testbed for testing different grasping strategies.

\subsection{Grasping Prediction}

Following previous work~\cite{pinto2016supersizing}, we formulate grasping prediction as a classification problem. Given a 2D input image $I$, taken by a camera with a top-down view, we sample $N_g$ image patches. We then discretize the space of grasp angles to $N_a$ different angles. We use the patches as input to a convolutional neural network, which predicts the probability of success for every grasping angle with the grasp location being the center of the patch. The output of the ConvNet is a $N_a$-dimensional vector giving the likelihood of each angle. This results in a $N_g \times N_a$  grasp probability matrix. The policy then chooses the best patch and angle to execute the grasp. The robot's policy thus uses as input the image $I$, and as output the grasp location $(x_g, y_g)$, which is the center of the sampled patch, and the grasping angle $\theta_g$: $\pi^\robot:I \mapsto (x_g, y_g, \theta_g)$. 

\subsection{Adversarial Disturbance}
After the robot grasps an object successfully, the human can attempt to pull the object away from the robot's end-effector, by applying a force of fixed magnitude. The action space is discrete with 6 different actions, one for each direction: up/down, left/right, inwards/outwards. As a result of the applied force, the object either remains on the robot's hand, or it is dropped to the ground.

\subsection{Network Architecture}
\begin{figure}[!t]
\centering
\includegraphics[width=1.0\linewidth]{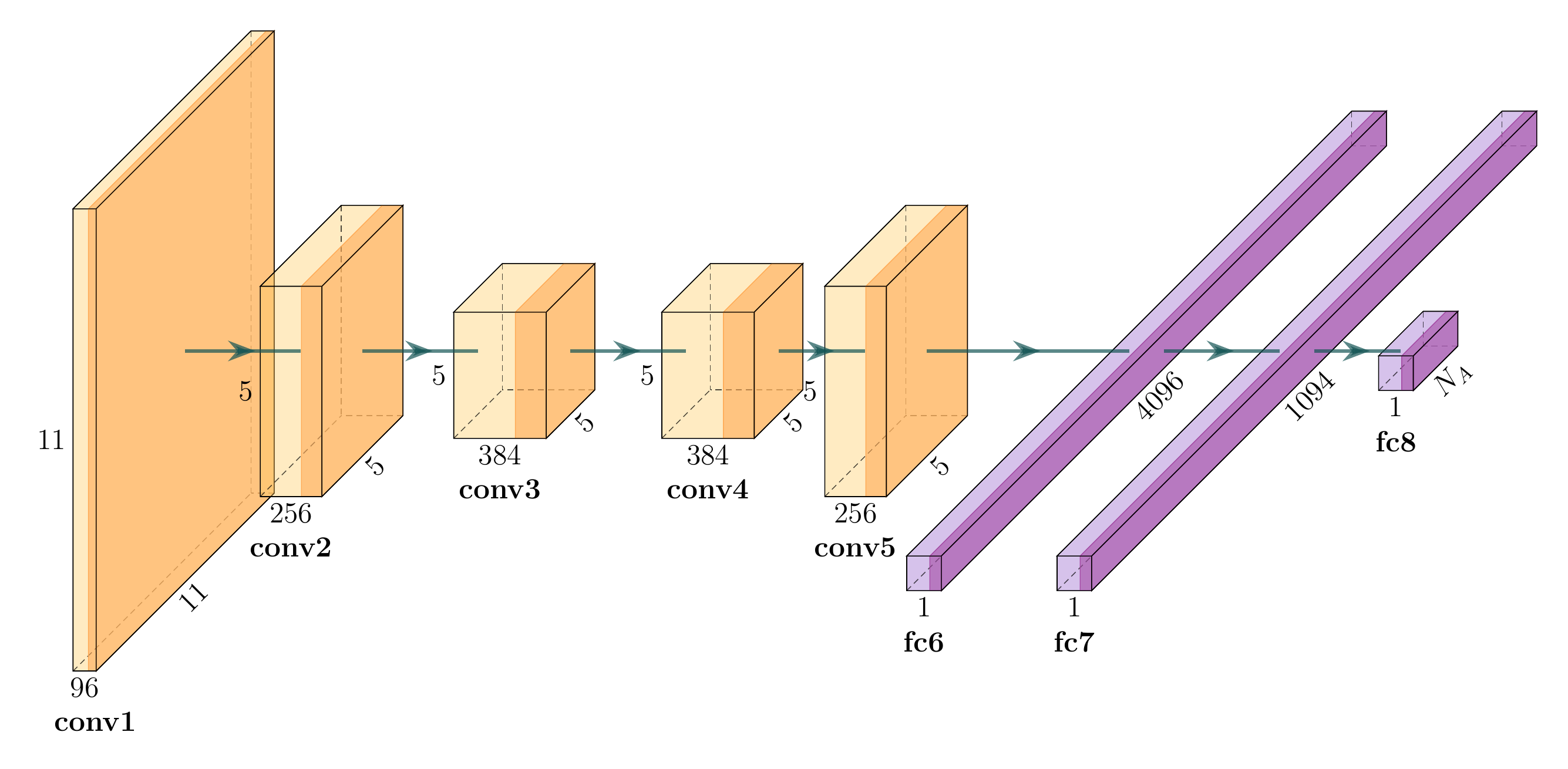}
\caption{ConvNet architecture for grasping.}
\label{fig:network}
\end{figure}

We use the same ConvNet architecture with previous work~\cite{pinto2016supersizing}, modeled on AlexNet~\cite{krizhevsky2012imagenet} and shown in Fig.~\ref{fig:network}. The output of the network is scaled to $(0,1)$ using a sigmoidal response function. 

\subsection{Network Training}
We initialized the network with a pretrained model initialized by Pinto et al.~\cite{pinto2016supersizing}. The model was pre-trained with completely different objects and patches. To train the model, we treat the reward $r$ that the robot receives as a training target for the network. Specifically, we set $R^{\robot}(s, a^\robot, s^+) = 1$ if the robot succeeds and 0 if the robot fails. Similarly, $R^{\human}(s^+, a^\human, s^{++})= 1$ if the human succeeds, and 0 if the human fails. Therefore, based on Eq.~\eref{eqn:reward}, the signal received by the robot is: 

\begin{equation}
r = \left\{\begin{matrix}
0 & \text{if robot fails to grasp}\\ 
1 & \text{if robot succeeds and human fails}\\ 
1 - \alpha & \text{if human succeeds}
\end{matrix}\right.
\label{eqn:target}
\end{equation}

We note that the training target is different than that of previous work~\cite{pinto2017supervision}.  There, the robot has access to the adversary's predictions by incorporating into the robot's loss function the probability that the adversarial network believes it can succeed. Here, however, the robot can only observe the outcome of the adversary's action.


We then define as loss function for the ConvNet, the binary cross entropy loss between the network's prediction and the reward received. We train the network using RMSProp~\cite{tieleman2012lecture}. 

\subsection{Simulation Environment}
For the training, we used the Mujoco~\cite{todorov2012mujoco} simulation environment. We customized the environment to allow a human user interacting with the physics engine.\footnote{The code is publicly available at: \url{https://github.com/icaros-usc/Interactive-mujoco_py}} 


\section{From Theory to Users}
We conducted a user study, with participants interacting with the robot in the virtual environment. The purpose of our study is to test whether the robustness of the robot's grasps can improve when interacting with a human adversary. We are also interested to explore how the object geometry affects the adversarial strategies of the users, as well as how users perceive robot's performance. 

\begin{figure}[!t]
\centering
\includegraphics[width=\linewidth]{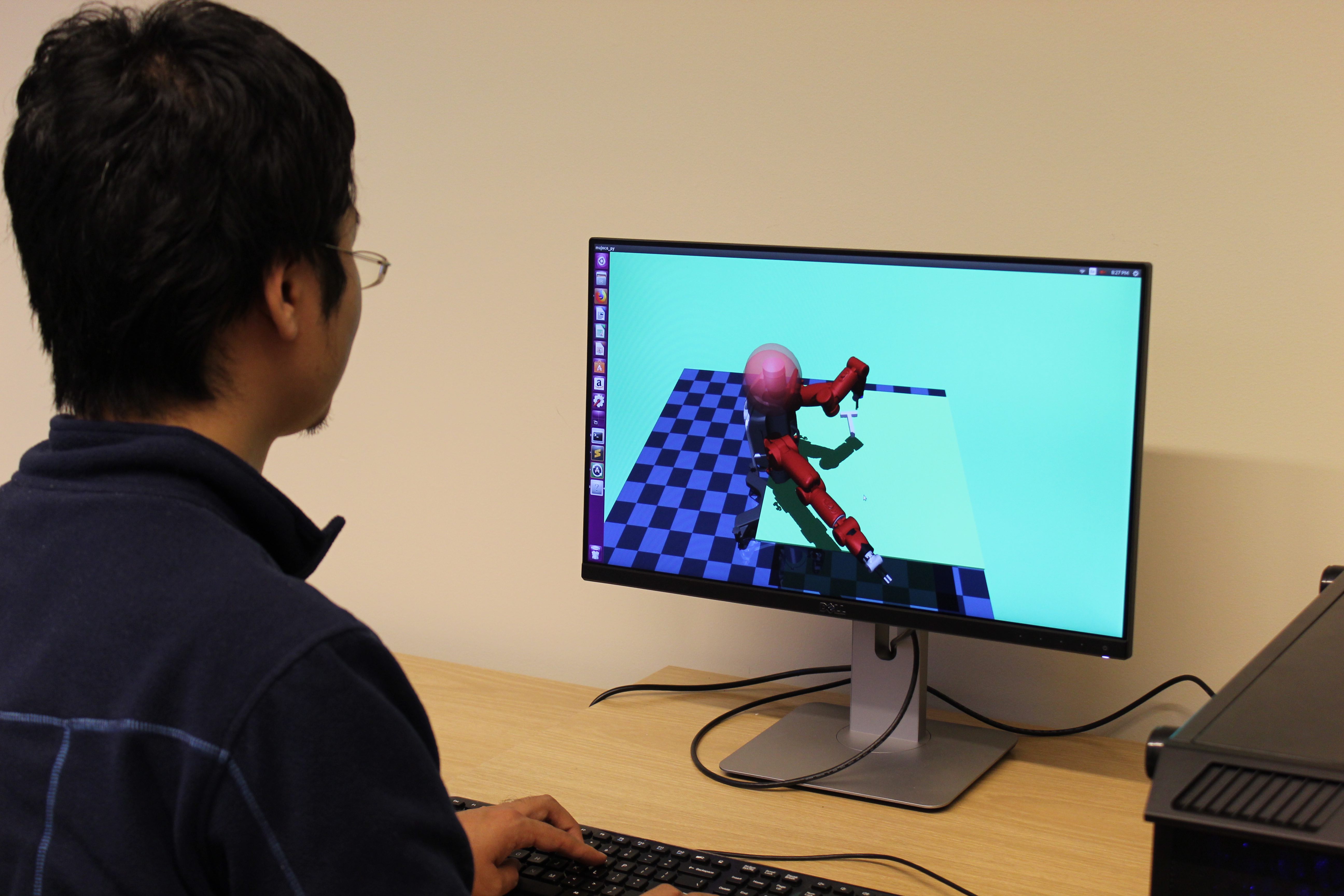}
\caption{Participants interacted with a simulated Baxter robot in the customized Mujoco simulation environment.}
\label{fig:experiment_setup}
\end{figure}

\noindent\textbf{Study Protocol.} Participants interacted with a simulated Baxter robot in the customized Mujoco simulation environment (Fig.~\ref{fig:experiment_setup}). The experimenter told participants that the goal of the study is to maximize robot's failure in grasping the object. They did not tell participants that the robot was learning from their actions. Participants applied forces to the object using the keyboard. All participants first did a short training phase by attempting to snatch an object from the robot's grasp 10 times, in order to get accustomed to the interface. The robot did not learn during that phase. Then, participants interacted with the robot executing Algorithm~\ref{alg:algorithm}. 

In order to keep the interactions with users short, we simplified the task, so that each user trained with the robot on one object only, presented to the robot at the same orientation. We fixed the magnitude of the forces applied to each object, so that the adversary would succeed if the grasp was unstable but fail to snatch the object otherwise. We selected a batch size $B=5$ and a number of episodes per batch $M=9$. The interaction with the robot lasted on average 10 minutes~\footnote{The anonymized log files of the human adversarial actions are publicly available at: \url{https://github.com/icaros-usc/human_adversarial_grasping_data}}.

\noindent\textbf{Manipulated variables.} We manipulated (1) the robot's learning framework and (2) the object that users interacted with. We had three conditions for the first independent variable: the robot interacting with a human adversary, the robot interacting with a simulated adversary that learns how to succeed in snatching the object and the robot learning in a self-supervised manner, without an adversary. Following previous work~\cite{pinto2017supervision}, the simulated adversary is trained with an identical network with training target equal to 1 if the snatching succeeds and 0 if the snatching fails.

We had five different objects (Fig.~\ref{fig:summary}). We selected objects of varying grasping difficulty and geometry to explore the different strategies employed by the human adversary. 


\noindent\textbf{Dependent measures.} 
For testing we executed the learned policy on the object for 50 episodes, applying a random disturbance after each grasp and recording the success or failure of the grasp before and after the random disturbance was applied. To avoid overfitting, we selected for testing the earliest learned model that met a selection criterion (early-stop)~\cite{caruana2001overfitting}. The testing was done using a script after the conduction of the study, without the participants being present. 
We additionally asked participants to report their agreement on a seven-point Likert scale to two statements regarding the robot's learning process (Table~\ref{tab:questionnaire}) and justify their answer.

\noindent\textbf{Hypotheses}\\
\noindent\textbf{H1}. \emph{We hypothesize that the robot trained with the human adversary will perform better than the robot trained in a self-supervised manner.}  
We base this hypothesis on previous work~\cite{pinto2017supervision} that has shown that training with a simulated adversary improved robot's performance, compared to training in a self-supervised manner.

\noindent\textbf{H2}.  \emph{We hypothesize that the robot trained with the human adversary will perform better than the robot trained with a simulated adversary.}
A human adversary has domain knowledge: they observe the object geometry and have intuition about the physics properties. Therefore, we expect the human to act as a \emph{model-based learning agent} and use their model to do targeted adversarial actions. On the other hand, the simulated adversary  has no such knowledge and they need to learn the outcome of different actions through interaction. 

\noindent\textbf{Subject allocation.} We recruited 25 users, 21 Male and 4 female participants. We followed a between-subjects design, where we had 5 users per object, in order to avoid confounding effects of humans learning to apply perturbations, getting tired or bored by the study. 

\begin{table}
    \centering
\caption{Likert Items.}
\label{tab:questionnaire}
\begin{tabular}{ l }

\hline
1. The robot learned throughout the study. \\
2. The performance of the robot improved throughout the study.\\
\hline
\end{tabular}
\end{table}

\newcommand{\cell}[2] {
\begin{tabular}{ll}#1 & #2 \end{tabular}
 }

\begin{table*}
    \centering
    \captionsetup{justification=centering,margin=2cm}
\caption{Grasping success rate (per cent) before (left column) and after (right column) application of random disturbance. Different users interacted with different objects (between-subjects design).}
\label{tab:results}
\begin{tabular}{r|ccccc}
User \# & Bottle & T-shape & Half-nut & Round-nut & Stick\\
\hline~\vspace{-5pt}\\
1 &\cell{~64}{~40} &\cell{~56}{~42} &\cell{~40}{~36} &\cell{~58}{~40} &\cell{~90}{~62}\\
2 &\cell{~64}{~40} &\cell{~52}{~28} &\cell{~40}{~36} &\cell{~82}{~48} &\cell{~94}{~64}\\
3 &\cell{~66}{~40} &\cell{~56}{~42} &\cell{~40}{~36} &\cell{~82}{~54} &\cell{~92}{~64}\\
4 &\cell{~74}{~40} &\cell{~78}{~60} &\cell{~40}{~36} &\cell{~52}{~40} &\cell{~90}{~62}\\
5 &\cell{~68}{~40} &\cell{~78}{~62} &\cell{~40}{~36} &\cell{~84}{~48} &\cell{100}{~84}\\
Simulated-adversary &\cell{~60}{~38} &\cell{~76}{~54} &\cell{~42}{~38} &\cell{54}{~50}&\cell{~64}{~54}\\
Self-trained &\cell{~14}{~~4}  &\cell{~52}{~34} &\cell{~40}{~36} &\cell{~80}{~40} &\cell{~50}{~18}\\
\end{tabular}
\end{table*}

\begin{figure*}[!t]
\centering
\includegraphics[width=1.0\linewidth]{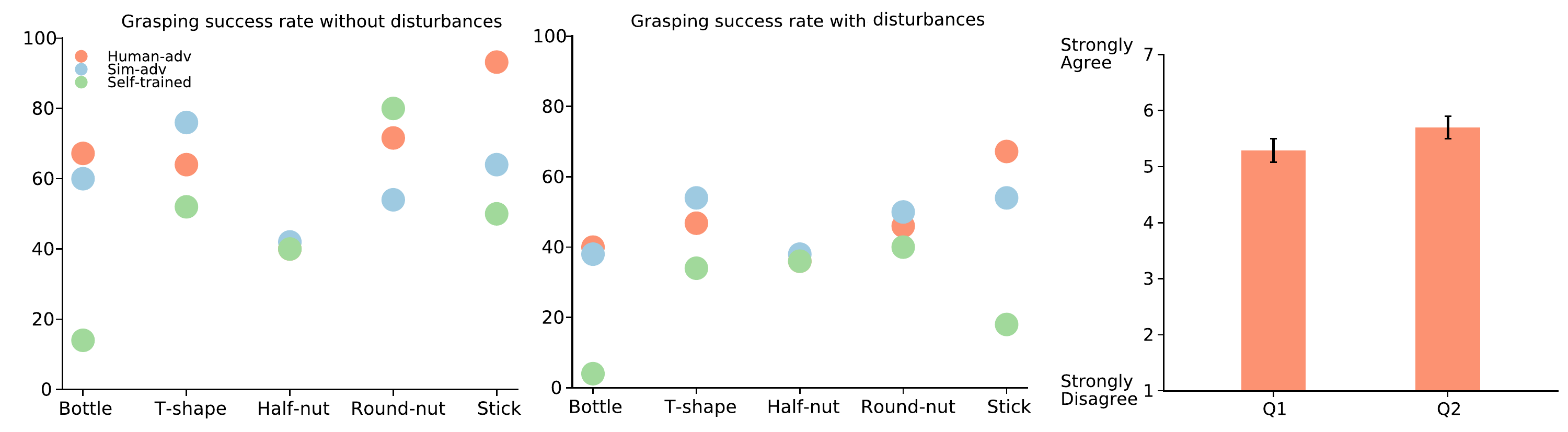}

\caption{Success rates from Table~\ref{tab:results} for all five participants  and subjective metrics.}
\label{fig:results}
\end{figure*}

\begin{figure*}[!t]
\centering
\begin{tabular}{ccccc}
\includegraphics[width=0.18\linewidth]{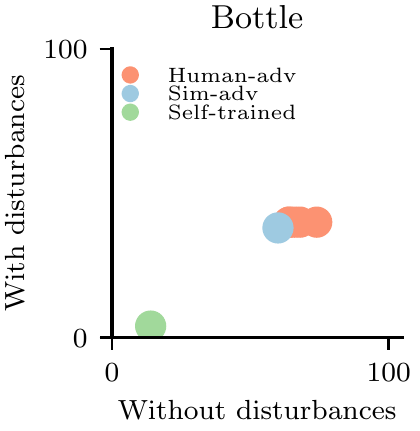}&
\includegraphics[width=0.18\linewidth]{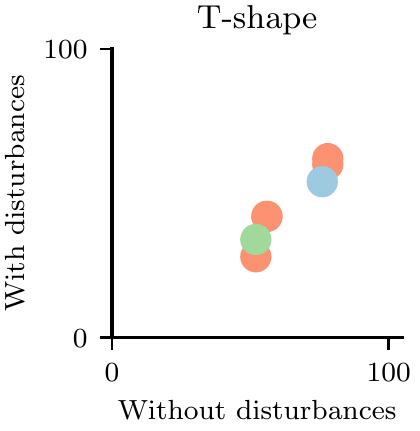}
\includegraphics[width=0.18\linewidth]{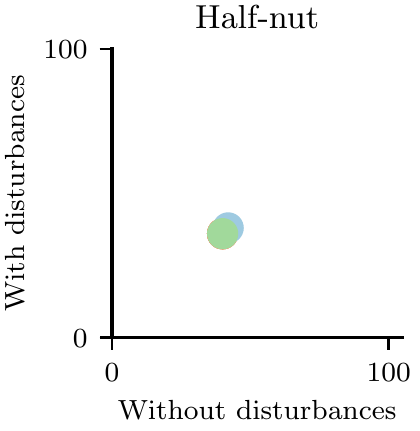}&
\includegraphics[width=0.18\linewidth]{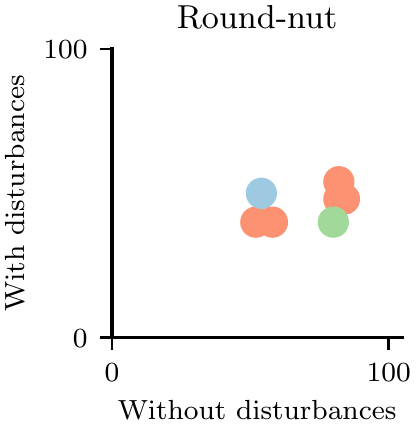}&
\includegraphics[width=0.18\linewidth]{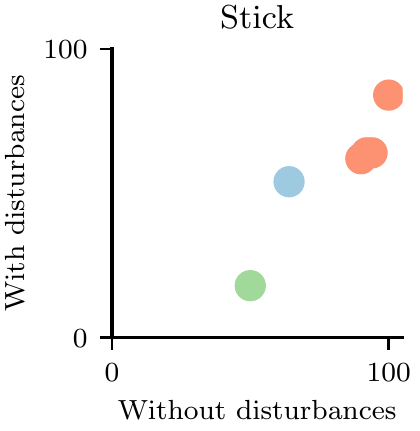}&
\end{tabular}
\caption{Success rates from Table~\ref{tab:results} for each object with (y-axis) and without (x-axis) random disturbances for all five participants.}
\label{fig:rates}
\end{figure*}

\begin{figure*}[!t]
\centering
\begin{tabular}{ccccc}
\begin{subfigure}[b]{0.18\linewidth}
\includegraphics[width=1.0\linewidth]{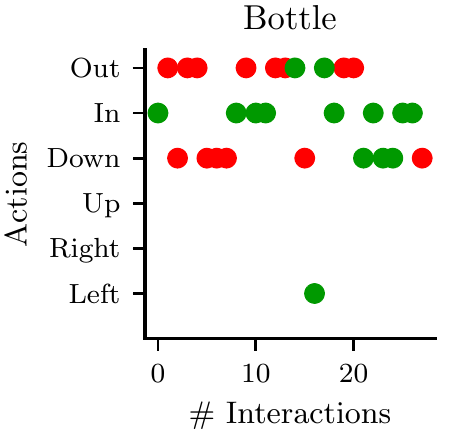}
\caption{}
\label{fig:a}
\end{subfigure}
&
\begin{subfigure}[b]{0.18\linewidth}
\includegraphics[width=1.0\linewidth]{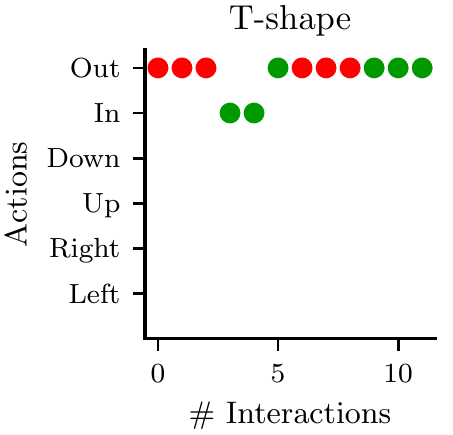}
\caption{}
\label{fig:b}
\end{subfigure}
&
\begin{subfigure}[b]{0.18\linewidth}
\includegraphics[width=1.0\linewidth]{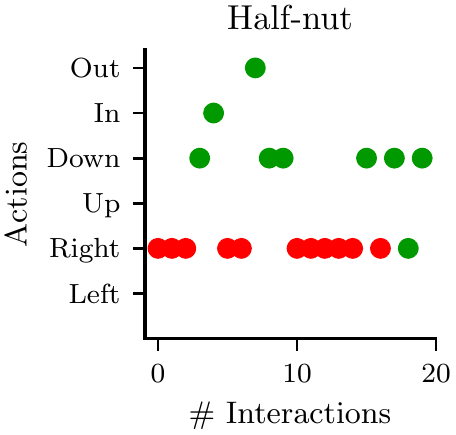}
\caption{}
\label{fig:c}
\end{subfigure}
&
\begin{subfigure}[b]{0.18\linewidth}
\includegraphics[width=1.0\linewidth]{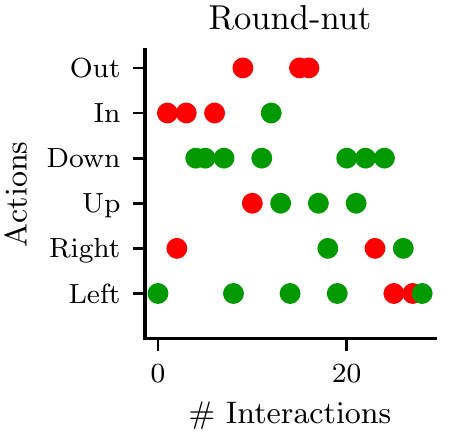}
\caption{}
\label{fig:d}
\end{subfigure}
&
\begin{subfigure}[b]{0.18\linewidth}
\includegraphics[width=1.0\linewidth]{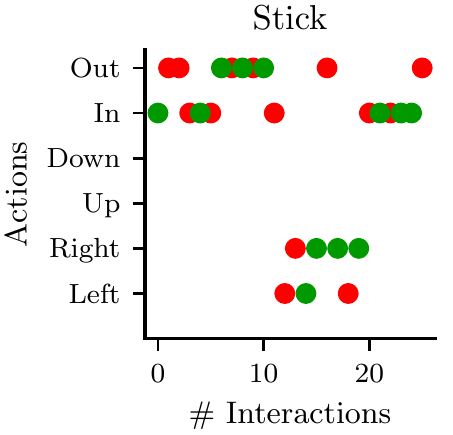}
\caption{}
\label{fig:e}
\end{subfigure}
\end{tabular}
\caption{Actions applied by selected human adversaries over time. We plot in green adversarial actions that the robot succeeds in resisting and in red actions that result in the human `snatching' the object.}
\label{fig:disturbances}
\end{figure*}

\section{Results}
\subsection{Analysis}
\noindent\textbf{Objective metrics.} Table~\ref{tab:results} shows the success rates for different objects. Different users interacted with each object; for instance User 1 for Bottle is a different participant than User 1 for T-shape. We have two dependent variables, the success rate of robot grasping an object in the testing phase in the absence of any perturbations, and the success rate with random perturbations being applied.
A two-way multivariate ANOVA \cite{kutner2005applied} with object and framework as independent variables showed a statistically significant interaction effect for both dependent measures: ($F(16,38) = 3.07, p = 0.002, \text{Wilks'~} \Lambda=0.19$).   In line with \textbf{H1}, a Post-hoc Tukey tests with Bonferroni correction showed that success rates were significantly larger for the human adversary condition than the self trained condition, both with ($p <0.001$) and without random disturbances ($p=0.001$).

We note that the post-hoc analysis should be viewed with caution, because of the significant interaction effect. To interpret these results, we plot the mean success rates for all conditions (Fig.~\ref{fig:results}). For clarity, we also contrast both success rates for each object separately in Fig.~\ref{fig:rates}.
 Indeed, we see the the success rate averaged over all human adversaries was higher for three out of five objects. The difference was largest for the bottle and the stick. The reason is that it was easy for the self-trained policy to pick up these objects without a robust grasp, which resulted in slow learning. On the other hand, the network trained with the human adversary rejected these unstable grasps, and learned quickly robust grasps for these objects. In contrast, round nut and half-nut objects could be grasped robustly at the curved areas of the object. The self-trained network thus got ``lucky'' finding these grasps, and the difference was negligible. In summary, these results lead to the following insight: 

\begin{quote}
Training with a human adversary is particularly beneficial for objects that have few robust grasp candidates that the network needs to search for.
\end{quote}

There were no significant differences between the rates in the human adversary and simulated adversary condition. Indeed, we see that the mean success rates were quite close for the two conditions. We expected the human adversary to perform better, since we hypothesized that the human adversary has a \emph{model} of the environment, which the simulated adversary does not have. Therefore, we expected the human adversarial actions to be more targeted. To explain this result,
which does not support \textbf{H2}, we look at human behaviors below.

\noindent\textbf{Behaviors.} Fig.~\ref{fig:disturbances} shows the disturbances applied over time for different users. Observing the participants behaviors, we see that some participants \emph{used their model of the environment to apply disturbances effectively}. Specifically, the user in Fig.~\ref{fig:b} applied a force outwards in the T-shape, succeeding in `snatching' the object even at the first try, which is indicated by the red dots. Gradually, the robot learned a more robust grasping policy, which resulted in the user failing to snatch the object (green dots). Similarly, the user in Fig.~\ref{fig:a} and Fig.~\ref{fig:c} used targeted perturbations which resulted in failed grasps from the very start of the task.


In some cases, such as in Fig.~\ref{fig:e},  the user adapted their strategy as well: when the robot learned to withstand an adversarial action outwards, the user acted by applying a force to the right, until the robot learned that as well. 

\begin{figure}[!t]
\centering
\begin{tabular}{cc}
\includegraphics[width=0.44\linewidth]{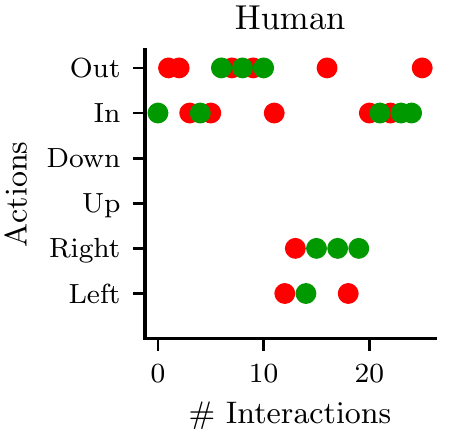}
\includegraphics[width=0.44\linewidth]{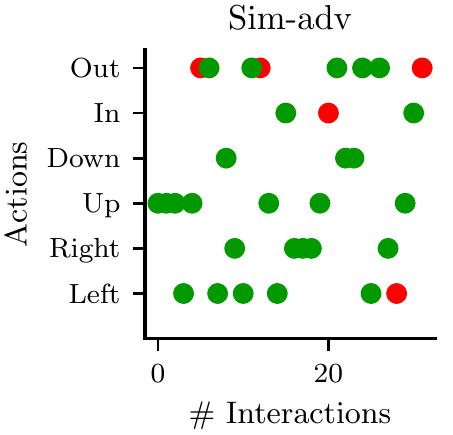}

\end{tabular}
\caption{Difference between training with user and simulated adversary for the stick object. The simulated adversary explores by applying forces in directions that fail to snatch the object. The red dots indicate human success in snatching the object, while the green dots indicate robot success in withstanding the human perturbation.}

\label{fig:comparison}
\end{figure}

Fig.~\ref{fig:comparison} compares the user of Fig.~\ref{fig:e} with the simulated adversary for the same object (stick). We observe that the simulated adversary explores different perturbations that are unsuccessful in snatching the object. This translates to worse performance for that object in the testing phase.

\begin{figure}[!t]
\centering
\begin{tabular}{cc}
\begin{subfigure}[b]{0.4\linewidth}
\includegraphics[width=1.0\linewidth]{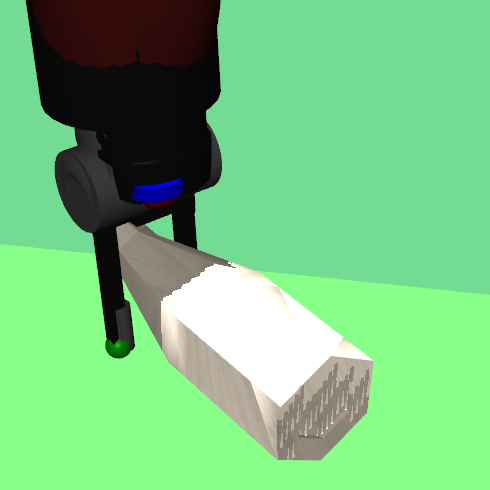}
\caption{}
\label{fig:unstable-bottle}
\end{subfigure} & 
\begin{subfigure}[b]{0.4\linewidth}
\includegraphics[width=1.0\linewidth]{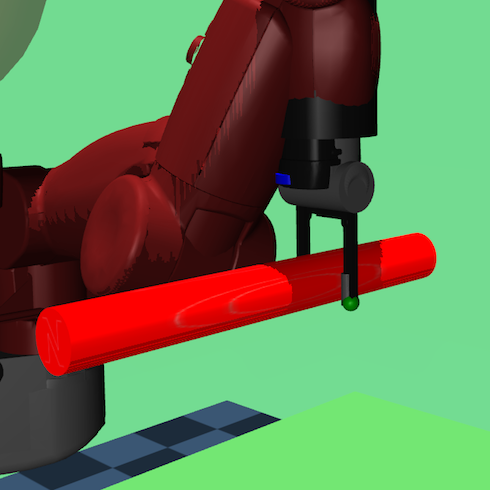}
\caption{}
\label{fig:unstable-stick}
\end{subfigure}
\end{tabular}
\caption{A force in almost any direction would make the grasp (a) fail, while only a force parallel to the axis of the stick would snatch the object in grasp (b).}

\label{fig:unstable}
\end{figure}

However, not all grasps required an informed adversary for the grasp to fail. For instance, for the grasped bottle in Fig.~\ref{fig:unstable-bottle}, there were many different directions where an applied force could succeed in removing the object. Therefore, having a model of the environment did not offer a significant benefit, since almost any disturbance would succeed in dropping the object. On the contrary, several grasps of the stick object failed only with targeted disturbances in the direction parallel to the object's major axis (Fig.~\ref{fig:unstable-stick}), which explains the difference in performance between human and simulated adversaries for that object.


\begin{figure}[!h]
\centering
\includegraphics[width=0.7\linewidth]{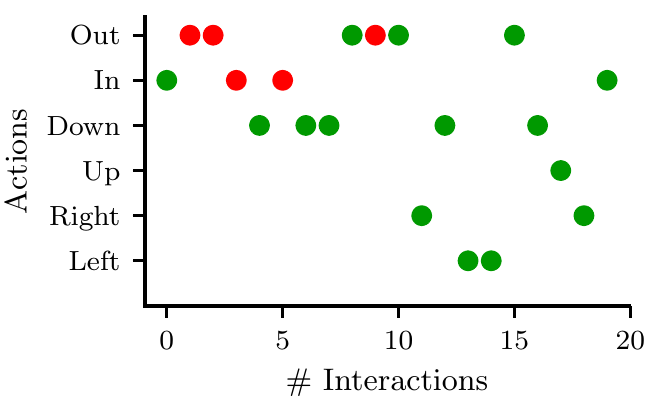}
\caption{The user started assisting the robot in the later part of the interaction, instead of acting as an adversary. The red dots indicate human success in snatching the object, while the green dots indicate robot success in withstanding the human perturbation.}
\label{fig:explore}
\end{figure}

Additionally, we found that some participants did not act as rational, model-based agents, which is the second factor that we believe affected the results. For instance, looking at one of the participants' interactions with the stick object (Fig.~\ref{fig:explore}), we see the variance of the actions increasing over time. We found this variance surprising, given the geometry of the object and the fact that all subsequent perturbations were unsuccessful. Looking at the open-ended responses, the participant stated that ``it seems some perturbations were challenging; so after some time I didn't apply that perturbation again.'' This indicates that at least one participant did not follow our instructions to act in an adversarial manner, and wanted to assist the robot instead.

\noindent\textbf{Subjective metrics.} We conclude our analysis with reporting the users' subjective responses (Fig.~\ref{fig:results}). A Cronbach's $\alpha=0.86$ showed good internal consistency~\cite{bland1997statistics}. Participants generally agreed that the robot learned throughout the study, and that its performance improved. In their open-ended responses, participants stated that ``The robot learned the grasping technique to win over me by learning from the forces that I provided and became more robust,'' and that ``The robot took almost 8 to 10 runs before it would start responding well. By the end of my experiment, it would grasp almost all the time.'' At the same time, one participant stated that the ``rate of improvement seemed pretty slow,'' and another that it ``kept making mistakes even towards the end.'' 

\subsection{Multiple objects.} 

We wish to test whether our framework can leverage human adversarial actions to learn grasping multiple objects at the same training session. Therefore, we modified the experiment setup, so that in each episode one of the five objects appeared randomly. To increase task difficulty, we additionally randomized the object's position and orientation in every episode. The robot then trained with one of the authors of the paper for 200 episodes. We then tested the trained model for another 200 episodes with randomly selected objects of random positions and orientations, as well as randomly applied disturbances. The trained model achieved a $52\%$ grasping success rate without disturbances, and $34\%$ success rate with disturbances. The rates were higher than those of a simulated adversary trained in the same environment for the same number of episodes, which had $28\%$ grasping success rate without disturbances and $22\%$ with disturbances. We find this result promising, since it indicates that targeted perturbations from a human expert can improve the efficiency and robustness of robot grasping. 
\section{Conclusion}
\noindent\textbf{Limitations.} Our work is limited in many ways. Our experiment was conducted in a virtual environment, and the users' adversarial actions were constrained by the interface. Our environment provides a testbed for different human-robot interaction algorithms in manipulation tasks, but we are also interested in exploring what types of adversarial actions users apply in real-world settings. We also focused on interactions with only one human adversary; a robot ``in the wild'' is likely to interact with multiple users. Previous work~\cite{pinto2017supervision} has shown that training a model with different robotic adversaries further improves performance, and it is worth exploring whether the same holds for human adversaries.

\noindent\textbf{Implications.} Humans are not always going to act cooperatively with their robotic counterparts. This work shows that from a learning perspective, this is not necessarily a bad thing. We believe that we have only scratched the surface of the potential applications of learning via adversarial human games: Humans can understand stability and robustness better than learned adversaries, and we are excited to explore human-in-the-loop adversarial learning in other tasks as well, such as obstacle avoidance for manipulators and mobile robots.



\bibliography{references}
\bibliographystyle{IEEEtran}
\end{document}